\pdfoutput=1
\documentclass[conference]{IEEEtran}
\IEEEoverridecommandlockouts

\usepackage[export]{adjustbox}
\usepackage{etex}
\usepackage{subfigure}
\usepackage{graphicx}
\usepackage{amsmath}
\usepackage{bm}
\usepackage{url}
\usepackage{stmaryrd}
\usepackage{balance}
\usepackage{amssymb}
\usepackage{pifont}
\usepackage[usenames,dvipsnames]{pstricks}
\usepackage{epsfig}
\usepackage{epstopdf}
\usepackage{multirow}
\usepackage{booktabs}
\usepackage{pgffor}
\usepackage{tikz}
\usepackage{eqparbox}
\usepackage{enumitem}
\usepackage{algorithm,algorithmic}
\usepackage[square,sort,comma,numbers]{natbib}

\newcommand{\pro}[0]{\textsc{HitFraud}}
\newcommand{\fullname}[0]{\textsc{Electronic Arts}}
\newcommand{\abbrname}[0]{\textsc{EA}}

\newtheorem{Definition}{\textsc{Definition}}
\newtheorem{Lemma}{\textsc{Lemma}}
\newtheorem{Proof}{\textsc{Proof}}

\begin{document}
\title{
HitFraud: A Broad Learning Approach for Collective Fraud Detection in Heterogeneous Information Networks\\
\thanks{This work was partially done while the author was an intern at Electronic Arts.}
}
\author{\IEEEauthorblockN{
Bokai Cao\IEEEauthorrefmark{1},
Mia Mao\IEEEauthorrefmark{2},
Siim Viidu\IEEEauthorrefmark{2} and
Philip S. Yu\IEEEauthorrefmark{1}\IEEEauthorrefmark{3}}
\IEEEauthorblockA{\IEEEauthorrefmark{1}Department of Computer Science, University of Illinois at Chicago, IL, USA; \{caobokai, psyu\}@uic.edu}
\IEEEauthorblockA{\IEEEauthorrefmark{2}Electronic Arts, Redwood City, CA, USA; \{mmao, sviidu\}@ea.com}
\IEEEauthorblockA{\IEEEauthorrefmark{3}Institute for Data Science, Tsinghua University, Beijing, China}}

\maketitle

\begin{abstract}
On electronic game platforms, different payment transactions have different levels of risk. Risk is generally higher for digital goods in e-commerce. However, it differs based on product and its popularity, the offer type (packaged game, virtual currency to a game or subscription service), storefront and geography. Existing fraud policies and models make decisions independently for each transaction based on transaction attributes, payment velocities, user characteristics, and other relevant information. However, suspicious transactions may still evade detection and hence we propose a broad learning approach leveraging a graph based perspective to uncover relationships among suspicious transactions, {\em i.e.}, inter-transaction dependency. Our focus is to detect suspicious transactions by capturing common fraudulent behaviors that would not be considered suspicious when being considered in isolation. In this paper, we present {\pro} that leverages heterogeneous information networks for collective fraud detection by exploring correlated and fast evolving fraudulent behaviors. First, a heterogeneous information network is designed to link entities of interest in the transaction database via different semantics. Then, graph based features are efficiently discovered from the network exploiting the concept of meta-paths, and decisions on frauds are made collectively on test instances. Experiments on real-world payment transaction data from {\fullname} demonstrate that the prediction performance is effectively boosted by {\pro} with fast convergence where the computation of meta-path based features is largely optimized. Notably, recall can be improved up to 7.93\% and F-score 4.62\% compared to baselines.
\end{abstract}

\begin{IEEEkeywords}
collective fraud detection, inter-transaction dependency, heterogeneous information network.
\end{IEEEkeywords}

\section{Introduction}
\label{sec:intro}

Fraud detection has attracted significant research efforts in recent years for various tasks including finance, security and web services. In this work, we investigate the fraud detection problem on electronic game platforms where it is desirable to identify suspicious payment transactions in an early stage in order to avoid chargebacks and to enhance normal users' experience. Current fraud models make independent decisions for each transaction, and detection becomes harder when intelligent adversaries are used, {\em e.g.}, proxy IP addresses. Graph based methods can detect frauds by leveraging the linkage information between entities of interest \cite{gyongyi2004combating,hooi2016fraudar,jiang2014catchsync}. Such methods are relatively harder to evade because making a fraud payment transaction unavoidably generates links in the graph which reveals {\em inter-transaction dependency}.

\begin{figure}[t]
\centering
    \begin{minipage}[l]{1.0\columnwidth}
      \centering
      \includegraphics[width=1\textwidth]{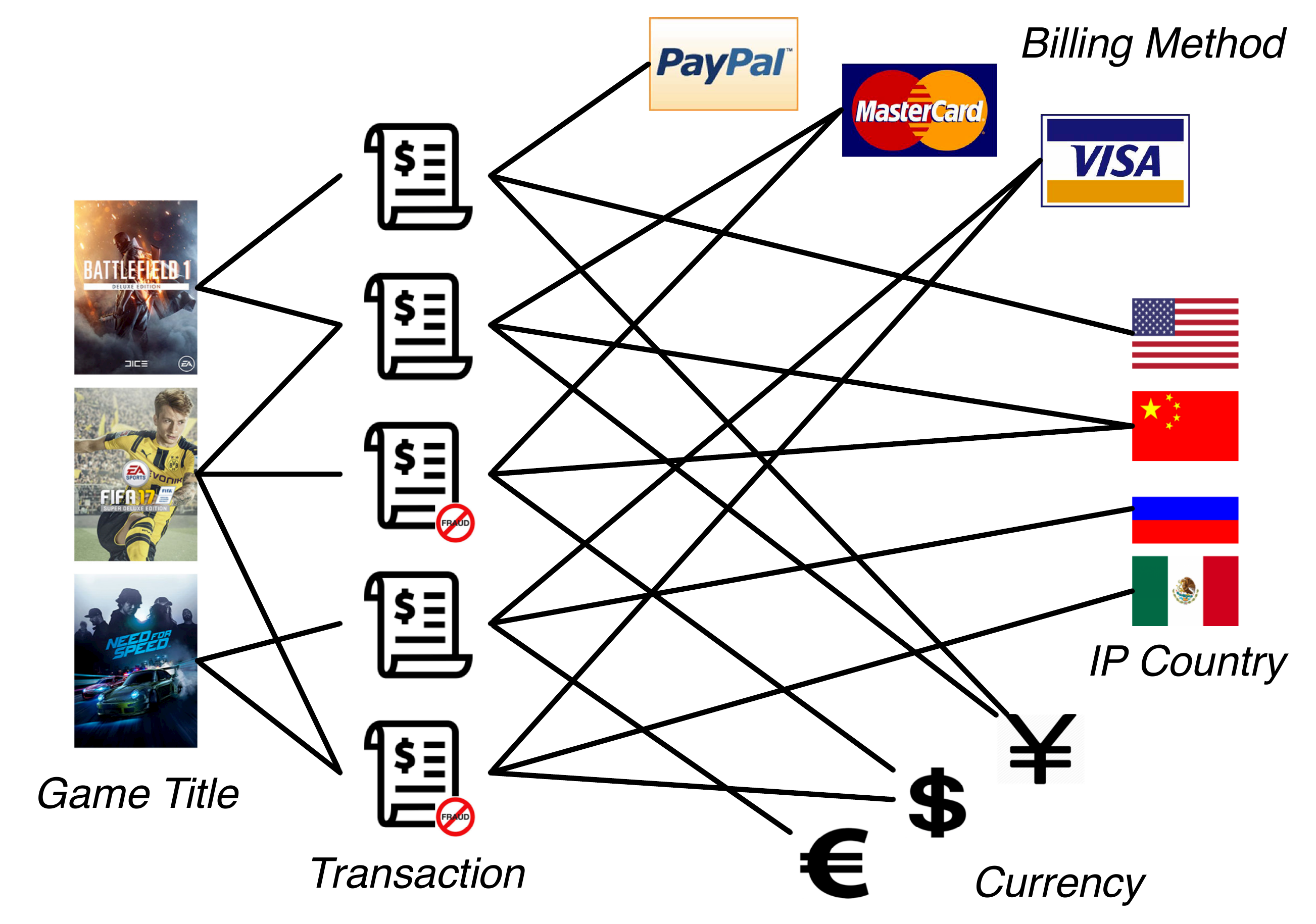}
    \end{minipage}
  \caption{A simplified example of the constructed heterogeneous information network.}\label{fig:intro}
\end{figure}

Existing graph based fraud detection approaches heavily focus on homogeneous information networks and bipartite graphs. Heterogeneous information networks (HINs) \cite{sun2011pathsim} are a special type of information networks that involve multiple types of nodes or multiple types of links, as shown in Figure~\ref{fig:intro}. In a HIN, different types of nodes and links have different semantic meanings. Such complex and semantically enriched networks possess great potential for knowledge discovery \cite{cao2015inferring,ji2011ranking,kong2013multi,sun2009ranking}. Its applications to fraud detection, however, are largely unexplored. Machine learning approaches that leverage a variety of data sources or fuse heterogeneous information can be referred to as {\em broad learning}.

Therefore, we are motivated to investigate how to leverage HINs to facilitate the fraud detection task. Most importantly, we seek to capture the inter-transaction dependency. It is critical to explore such relationships among suspicious transactions because fraudulent behaviors are often {\em correlated} and {\em fast evolving}. (1) HINs provide us with an effective and compact representation of linked transactions in various semantics, {\em e.g.}, the same currency, the same IP address, and the same game titles. The statistics of the label information ({\em i.e.}, fraud or normal) of these linked transactions can be aggregated, and thereby add a new dimension of measurements to distinguish suspicious transactions from normal ones based on the correlated fraudulent behaviors. (2) In order to tackle the problem of fast evolving fraudulent behaviors, we should not only consider the inter-transaction dependency across training transactions and test transactions, but also include the dependency among test transactions. Hence, suspicious transactions are identified in a semi-supervised manner by iteratively obtaining the predicted labels of test transactions and updating the statistics of linked transactions in alternation. Such a collective prediction procedure has the potential to detect a suspicious transaction even if it appears to be normal by itself but its linked transactions (other test transactions in a batch sharing categorical variables) are identified as very suspicious, and thereby improve the recall metric. The main contributions of this work are fourfold:
\begin{itemize}[leftmargin=*]
\setlength\itemsep{0pt}

\item As fraud payment transactions generally do not occur in isolation, {\em i.e.}, fraudulent behaviors are often correlated and fast evolving, we formulate the fraud detection task as a collective prediction problem in a HIN to capture relationships among fraud payment transactions. (Section~\ref{sec:problem})

\item To address the daunting challenge on huge feature space, we design a HIN that can effectively capture the various relationships among transactions. Here meta-paths (a sequence of link types) are explored to identify the relevant inter-transaction dependency features. (Section~\ref{sec:dataset})

\item We propose an effective and efficient algorithm to compute meta-path based features in the framework of collective fraud detection. (Section~\ref{sec:method})

\item We validate that the observation that fraudulent behaviors are often correlated and fast evolving can indeed be explored to more effectively capture fraud payment transactions. We evaluate the proposed framework on real-world payment transaction data from {\fullname} payment system, and results show that recall and F-score can significantly be improved by exploring inter-transaction dependency via the proposed collective fraud detection method based on HINs. (Section~\ref{sec:exp})

\end{itemize}

\section{Problem Definition}
\label{sec:problem}

\noindent\textbf{Heterogeneous Information Network (HIN).}
A heterogeneous information network is a type of information network with multiple types of nodes or multiple types of links \cite{sun2011pathsim,sun2009ranking}. It can be represented as a directed graph $\mathcal{G}=\left(\mathcal{V},\mathcal{E}\right)$. $\mathcal{V} = \mathcal{V}^1 \cup \cdots \cup \mathcal{V}^m$ denotes the set of nodes involving $m$ node types: $\mathcal{V}^1= \{v^1_{1}, \cdots, v^1_{n_1}\}, \cdots , \mathcal{V}^m= \{v^m_{1}, \cdots, v^m_{n_m}\}$ where $v^i_{p}$ represents the $p$-th node of type $i$. $\mathcal{E} = \mathcal{E}^1 \cup \cdots \cup \mathcal{E}^r \subseteq \mathcal{V} \times \mathcal{V}$ denotes the set of links between nodes in $\mathcal{V}$ involving $r$ link types. Mathematically, a link type $k$ starting from source nodes of type $i$ and ending at target nodes of type $j$ is described by an adjacency matrix $\mathbf{A}^k \in \mathbb{R}^{n_i \times n_j}$ where $\mathbf{A}^k[p,q]=1$ if there exists a link in $\mathcal{E}^k$ between $v^i_p$ and $v^j_q$, otherwise $\mathbf{A}^k[p,q]=0$. We can write this link type as ``$\mathcal{V}^i \xrightarrow{\mathcal{E}^k} \mathcal{V}^j$".

\noindent\textbf{Collective Fraud Detection.}
We may assume without loss of generality that nodes in $\mathcal{V}^1$ are the target entities, {\em i.e.}, transactions, in the case of fraud detection. The number of target entities is denoted as $n=n_1$. Formally, we are given a data matrix $\mathbf{X} = [\mathbf{x}_1^T; \cdots; \mathbf{x}_{n}^T]$ where $\mathbf{x}_i \in \mathbb{R}^d$ is the feature vector of the $i$-th transaction which typically involves extensive feature engineering and domain knowledge. Labels are denoted as $\mathbf{y} = [y_1,\cdots,y_{n}]$ where $y_i=1$ if the $i$-th transaction is a fraud, otherwise $y_i=0$. The target is to identify the label $y_u$ of an unseen transaction $\mathbf{x}_u$ (typically, with different items, user accounts, billing accounts, and IP addresses, {\em etc.}).

In this paper, we investigate how to leverage a HIN to facilitate the fraud detection task. The problem of collective fraud detection in a HIN reduces to learning a predictive function $f:(\mathcal{V},\mathcal{E},\mathbf{X})\to\mathbf{y}$. It is a nontrivial problem due to threefold challenges:
\begin{itemize}[leftmargin=*]
\setlength\itemsep{0pt}
\item We need to design a network that can capture various relationships among transactions.
\item We need to discover relevant network features efficiently, whose time complexity should ideally be linear to the number of transactions and insensitive to the type of relationships.
\item We need to effectively explore dependency among suspicious transactions, especially among the test ones, in order to detect the correlated and fast evolving fraudulent behaviors.
\end{itemize}


\begin{figure}[t]
\centering
    \begin{minipage}[l]{1.0\columnwidth}
      \centering
      \includegraphics[width=1\textwidth]{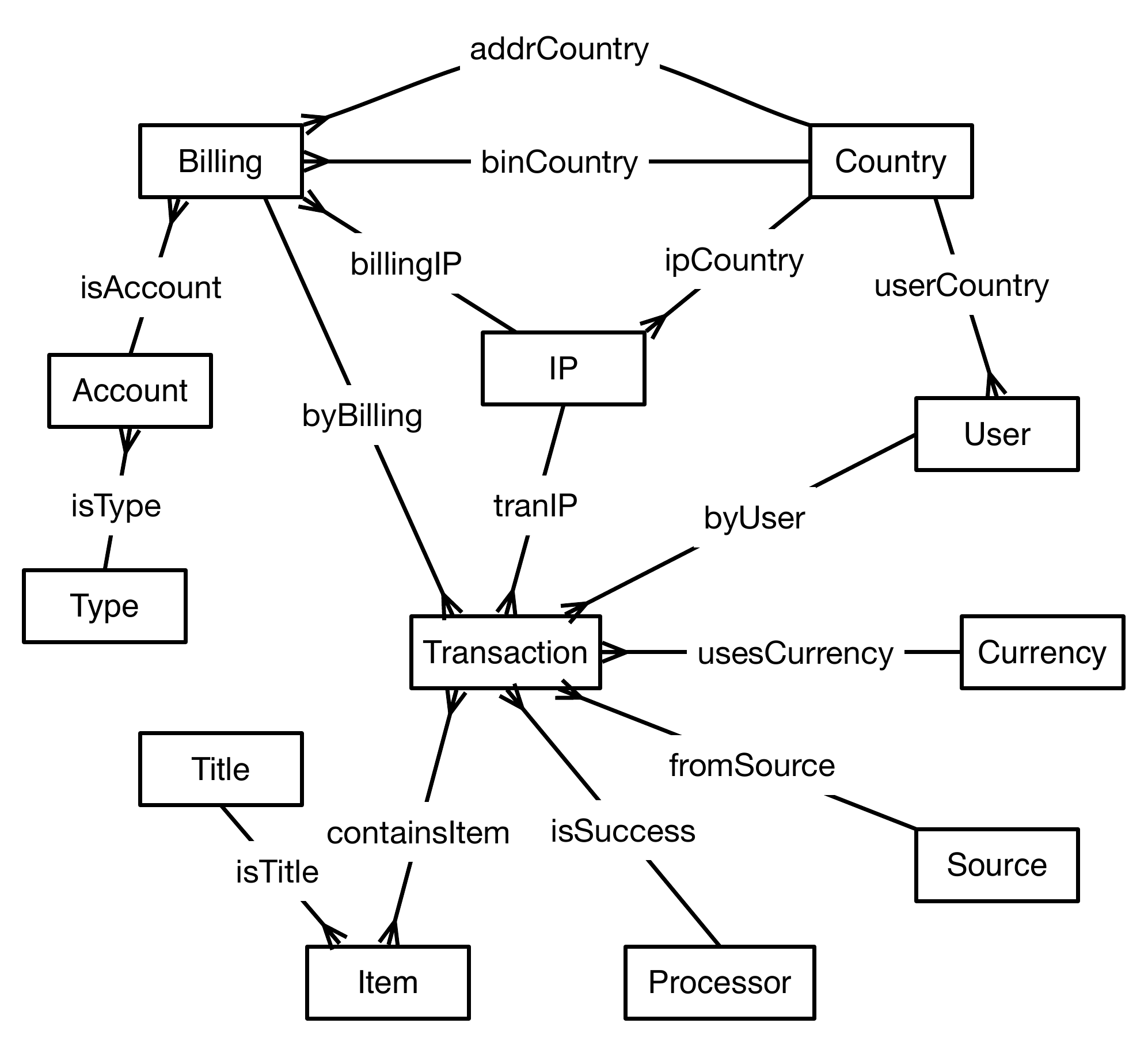}
    \end{minipage}
  \caption{The network schema of {\abbrname} payment transaction data. Each rectangle represents a node type, and each line represents a link type.}\label{fig:schema}
\end{figure}

\begin{table*}[t]
\caption{Top-10 discriminative meta-paths and their semantics.}
\label{tab:metapath}
\centering
\begin{tabular}{||l|l|l||}
\hline
ID & Meta-path & Semantics\\
\hline
\hline
\#1 & transaction $\xrightarrow{byBilling}$ billing $\xrightarrow{billingIP}$ IP $\xrightarrow{billingIP^{-1}}$ billing $\xrightarrow{byBilling^{-1}}$ transaction & from the same billing IP address\\
\#2 & transaction $\xrightarrow{tranIP}$ IP $\xrightarrow{tranIP^{-1}}$ transaction & from the same transaction IP address\\
\#3 & transaction $\xrightarrow{byBilling}$ billing $\xrightarrow{byBilling^{-1}}$ transaction & using the same billing account\\
\#4 & transaction $\xrightarrow{byUser}$ user $\xrightarrow{byUser^{-1}}$ transaction & using the same user account\\
\#5 & transaction $\xrightarrow{tranIP}$ IP $\xrightarrow{billingIP^{-1}}$ billing $\xrightarrow{byBilling^{-1}}$ transaction & transaction IP and billing IP are the same\\
\#6 & transaction $\xrightarrow{byBilling}$ billing $\xrightarrow{billingIP}$ IP $\xrightarrow{tranIP^{-1}}$ transaction & billing IP and transaction IP are the same\\
\#7 & transaction $\xrightarrow{containsItem}$ item $\xrightarrow{containsItem^{-1}}$ transaction & containing the same item\\
\#8 & transaction $\xrightarrow{containsItem}$ item $\xrightarrow{isTitle}$ title $\xrightarrow{isTitle^{-1}}$ item $\xrightarrow{containsItem^{-1}}$ transaction & containing the same game title\\
\#9 & transaction $\xrightarrow{fromSource}$ source $\xrightarrow{fromSource^{-1}}$ transaction & from the same source\\
\#10 & transaction $\xrightarrow{byUser}$ user $\xrightarrow{userCountry}$ country $\xrightarrow{userCountry^{-1}}$ user $\xrightarrow{byUser^{-1}}$ transaction & from the same user country\\
\hline
\end{tabular}
\end{table*}

\section{Dataset}
\label{sec:dataset}

In this work, we use {\fullname} ({\abbrname}) payment transaction data as an example to do the study. All transactions on {\abbrname} digital platform go through a set of policies, rules and models to determine their levels of risk, and a subset of them are sent for additional manual review. An experienced team reviews those transactions and decides if they should be rejected or approved, and this review decision is used as the ground-truth for training and evaluating fraud detection algorithms. We collected manual review data for $n=130K$ transactions during a recent period.
Each transaction is associated with a $d=2K$ dimensional feature vector, including transaction attributes, payment velocities, user characteristics, and other relevant information.

A HIN is constructed by linking entities of interest from several selected databases. Transactions are the target instances on which fraud decisions are made, so each transaction ID is represented as a node in the network, and the set of transaction IDs compose a node type in the network schema. In addition, other entities that are directly or indirectly related to a transaction are considered here, and they compose other node types in the schema, including billing accounts, user accounts, game titles, IP addresses, {\em etc.} Links are added based on common semantics. For example, a transaction is linked with a user if the user placed the transaction, and a transaction is linked with an item if the transaction contains the item.

As a result, the constructed HIN is composed of over $400K$ nodes and $1.5M$ links. It integrates data involving $m=12$ types of nodes, such as {\em transaction, user, item, title, currency, source, country, etc.} which are connected through $r=15$ types of links, such as ``transaction $\xrightarrow{containsItem}$ item", ``billing $\xrightarrow{billingIP}$ IP", {\em etc.} Its network schema is shown in Figure~\ref{fig:schema}. Note that {\em billing} and {\em country} are directly linked in two different semantics: country corresponding to the billing address, and country associated with the Bank Identification Number (BIN). Each billing account is designed to be associated with one country following each semantics.

\section{Method}
\label{sec:method}

In this paper, we present a broad learning approach, named {\pro}, for collective fraud detection in HINs by capturing inter-transaction dependency. The proposed framework is outlined as: 
\begin{enumerate}[leftmargin=*]
\setlength\itemsep{0pt}
\item The aforementioned HIN is designed so that it contains entities of interest as nodes, {\em e.g.}, transactions, billing accounts, user accounts, game titles, and IP addresses. Two entities are linked if they appear together in a record.
\item Efficient ways of finding downsized meta-paths from transaction nodes are explored using breadth-first-search on the network schema. Meta-path based features are computed by pairing the pre-computed downsized meta-paths.
\item The predicted labels of test transactions are obtained from a trained classifier, and meta-path based features are updated using both the labels of training transactions and the predicted labels of test transactions, in alternation until convergence.
\end{enumerate}

\subsection{Capturing Inter-Transaction Dependency}

First, we briefly review the concept of {\em meta-path} following previous work \cite{cao2014collective,kong2013multi,sun2011pathsim}. It has been demonstrated to be useful for mining P2P lending networks \cite{zhang2016netcycle}, social networks \cite{cao2015inferring}, bioinformatic networks \cite{cao2014collective,kong2013multi}, and bibliographic networks \cite{kong2012meta,sun2011pathsim}.

In general, a meta-path corresponds to a type of path within the network schema, containing a certain sequence of link types. For example, in Figure~\ref{fig:schema}, a meta-path ``transaction $\xrightarrow{containsItem}$ item $\xrightarrow{isTitle}$ title $\xrightarrow{isTitle^{-1}}$ item $\xrightarrow{containsItem^{-1}}$ transaction" denotes a composite relation between transactions where $containsItem^{-1}$ represents the inverted relation of \emph{containsItem}. The semantic meaning of this meta-path is that transactions contain items that belong to the same game title. Different meta-paths usually represent different semantic meanings between linked nodes. In this manner, various relationships among transactions can be described by a set of meta-paths. By capturing such inter-transaction dependency and aggregating the label information of the linked transactions, we could better detect correlated fraudulent behaviors. In other words, we could identify transactions with highly risky values in a categorical variable, {\em e.g.}, game title. Intuitively, when a recent launch of FIFA attracts many frauds to it, transactions that contain FIFA would be more suspicious than others. Similarly, transactions that are related to user accounts, billing accounts, IP addresses and currencies with high risk could also be identified. In Table~\ref{tab:metapath}, we show several examples of meta-paths that are used in this work and their semantics.

The implementation of meta-paths is essentially a chain of matrix multiplications. Let's denote a meta-path as $\mathcal{P}=<\mathcal{E}^{k_1},\cdots,\mathcal{E}^{k_l}>$ where the source node of $\mathcal{E}^{k_1}$ is of type $s$ and the target node of $\mathcal{E}^{k_l}$ is of type $t$. The semantic meaning of this meta-path is mathematically described as $\mathbf{P}=\mathbf{A}^{k_1}\times\cdots\times\mathbf{A}^{k_l} \in \mathbb{R}^{n_s \times n_t}$. Since $\mathbf{P}$ is usually asymmetric, the meta-paths \#5 and \#6 in Table~\ref{tab:metapath} are not identical. It is usually assumed that the strength of connection between $v^s_p$ and $v^t_q$ on such semantics is positively correlated with $\mathbf{P}[p,q]$, because $\mathbf{P}[p,q]$ is the (weighted) count of paths connecting $v^s_p$ and $v^t_q$ that follow the sequence of links in $\mathcal{P}$. Hereinafter, $\mathcal{P}$ and $\mathbf{P}$ will be used interchangeably when the meaning is clear from context.

\begin{figure}[t]
\centering
  \subfigure[Sparse links $\mathbf{A}$.]{
    \begin{minipage}[l]{0.35\columnwidth}
      \centering
      \includegraphics[height=4.5cm,center]{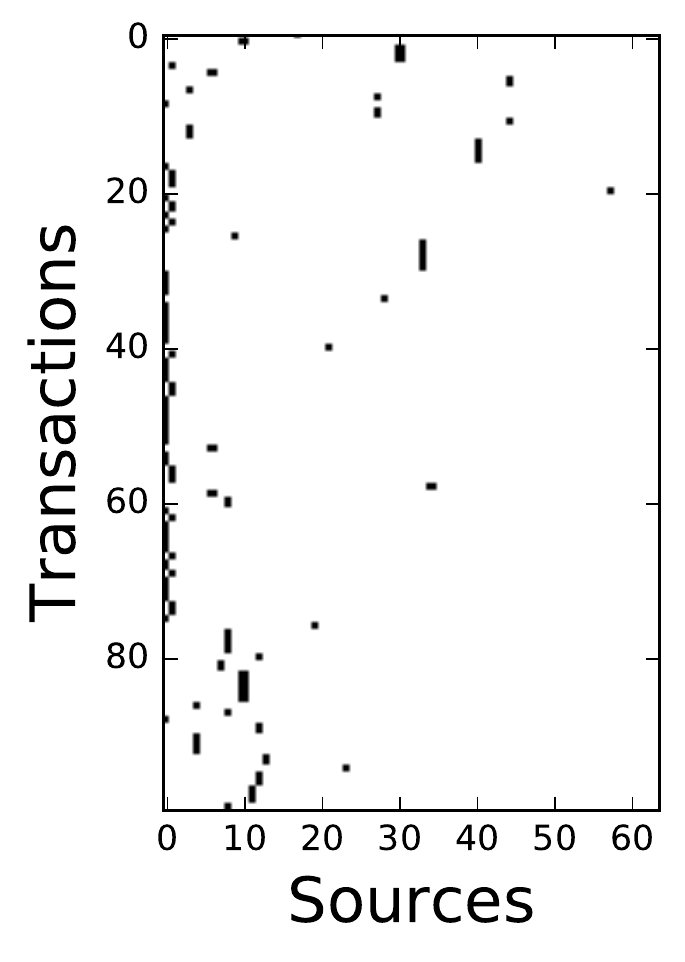}
    \end{minipage}
  }
  \subfigure[A meta-path $\mathbf{P}$.]{
    \begin{minipage}[l]{0.55\columnwidth}
      \centering
      \includegraphics[height=4.5cm,center]{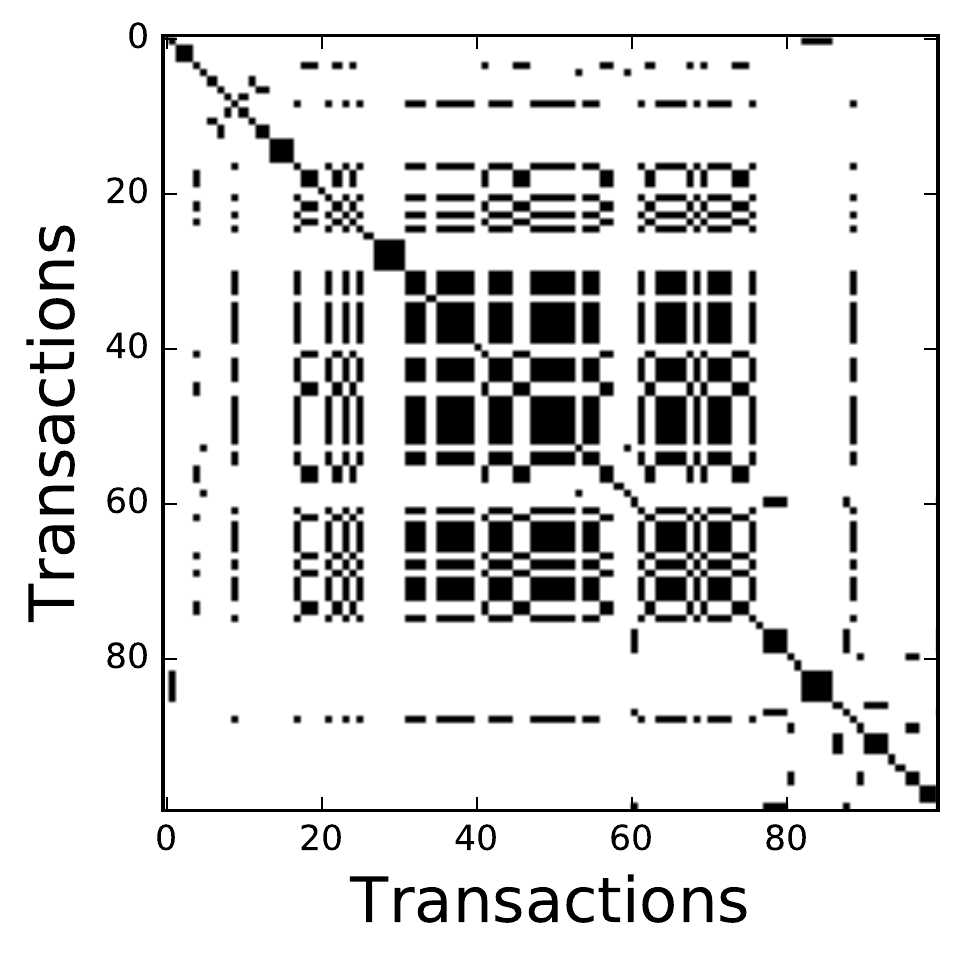}
    \end{minipage}
  }\
\caption{An example of computing a meta-path from sparse links.}\label{fig:example}
\end{figure}

\subsection{Redundancy in Meta-Paths}
\label{sec:redundancy}

Cardinality, which refers to the maximum number of times a node of the source node type can be linked with nodes of the target node type, is represented by the styling of a line and its endpoint in Figure~\ref{fig:schema}. The notation style is similar to entity-relationship diagrams (ERDs) where a Crow's foot shows many-to-one relationship.

Let's consider a meta-path ``transaction $\xrightarrow{fromSource}$ source $\xrightarrow{fromSource^{-1}}$ transaction" in Table~\ref{tab:metapath} and denote it as $\mathcal{P}=<\mathcal{E}^{k_1},\mathcal{E}^{k_2}>$ where $\mathcal{E}^{k_1}$ is the set of links indicating which source a transaction is from, and $\mathcal{E}^{k_2}$ is its inverted relation. The adjacency matrix of $\mathcal{E}^{k_1}$ is denoted as $\mathbf{A} \in \mathbb{R}^{n \times n_s}$ where $n=130K$ is the number of transactions and $n_s$ is the number of sources, then that of $\mathcal{E}^{k_2}$ is $\mathbf{A}^T$, and $\mathbf{P}=\mathbf{A}\times\mathbf{A}^T$. Note that an inverted relation is described by the transpose of its original adjacency matrix rather than the inverse. Because each transaction is conducted on one of {\abbrname} game stores, $\mathcal{E}^{k_1}$ here is a many-to-one relation between transactions and sources. That is to say, $\mathbf{A}$ is extremely sparse with one value per row, and its sparsity ratio is $1-1/n_s>98\%$.

The adjacency matrix $\mathbf{A}$ is shown in Figure~\ref{fig:example} for 100 randomly sampled transactions, as well as the the meta-path $\mathbf{P}=\mathbf{A}\times\mathbf{A}^T$. As we can see, through the direct application of matrix chain multiplication, the computation of a meta-path almost turns a sparse adjacency matrix $\mathbf{A}$ into a full matrix $\mathbf{P}$. Moreover, there are a lot of redundancy in $\mathbf{P}$. For this particular meta-path, if $i$-th transaction and the $j$-th transaction are from the same source $k$, {\em i.e.}, $\mathbf{A}[i,k]=\mathbf{A}[j,k]=1$, they have exactly the same row and column in $\mathbf{P}$, {\em i.e.}, $\mathbf{P}[i,:]=\mathbf{P}[j,:]=\mathbf{P}[:,i]=\mathbf{P}[:,j]$.

\subsection{Efficient Computation}
\label{sec:computation}

Multiplying adjacency matrices of a meta-path in the natural sequence can be inefficient from the time perspective, considering the classic matrix chain multiplication problem. It can be optimized using dynamic programming in $O(l^3)$ where $l$ is the length of a meta-path which is usually very small. In practice, however, the real difficulty in computing meta-paths lies in the space constraint. As we discussed above, turning sparse matrices into a full matrix is definitely not desirable because it makes the space cost from linear to quadratic with respect to the number of transactions.

Fortunately, we are not interested in the concrete form of a meta-path itself. For the purpose of obtaining features for fraud detection on transactions, the meta-paths that we need to compute should have the same source node type and target node type which is a transaction. Because each transaction may be linked with different number of transactions through a meta-path, aggregation functions are employed to combine the label information of linked transactions in order to derive a fixed number of meta-path based features. For example, we can use the weighted label fraction of linked transactions as the feature $\mathbf{z} \in \mathbb{R}^n$ for each meta-path \cite{kong2013multi,kong2012meta}. It is formulated as follows:
\begin{align}
\mathbf{z} = \mathbf{D} \times \mathbf{P} \times \mathbf{y}
\label{eq:relfea1}
\end{align}
where $\mathbf{D} \in \mathbb{R}^{n \times n}$ is a diagonal matrix and $\mathbf{D}[i,i] = 1/\sum_j \mathbf{P}[i,j]$.
In this manner, $z_i$ indicates the ratio of being frauds among transactions that are connected with the $i$-transaction through the meta-path.

\subsubsection{Many-to-One Cases}

Next, we will introduce how to obtain the feature $\mathbf{z}$ without explicitly computing the meta-path $\mathbf{P}$. Since the concept of meta-path is defined as a sequence of link types, it can also be considered as a sequence of node types that are endpoints of these links (different link sequences can correspond to the same node sequence though), say, $\mathcal{P}=<\mathcal{V}^{k_0},\cdots,\mathcal{V}^{k_l}>$. We have $k_0=k_l=1$ since we only consider meta-paths whose source node type and target node type are transaction. It is assumed that transaction is always the node set of the largest size, {\em i.e.}, $\operatornamewithlimits{argmax}_i\{|\mathcal{V}^i||\mathcal{V}^i\in\mathcal{P}\}=1$, and $t$ denotes the node set of the smallest size in $\mathcal{P}$, {\em i.e.}, $\operatornamewithlimits{argmin}_i\{|\mathcal{V}^i||\mathcal{V}^i\in\mathcal{P}\}=t$. Therefore, we can decompose a meta-path $\mathcal{P}$ into two parts at the node type $t$, and its matrix form can be written as $\mathbf{P} = \mathbf{P}_1 \times \mathbf{P}_2^T$ where $\mathbf{P}_1, \mathbf{P}_2 \in \mathbb{R}^{n \times n_t}$.

\begin{Definition}[Simple Meta-path]
A meta-path is a simple meta-path if it is a sequence of many-to-one relations.
\end{Definition}

Note that a simple meta-path itself is a (composite) many-to-one relation. For example, ``transaction $\xrightarrow{byBilling}$ billing $\xrightarrow{isAccount}$ account $\xrightarrow{isType}$ type" is a simple meta-path. All meta-paths in Table~\ref{tab:metapath} are a concatenation of a simple meta-path and another meta-path except for the seventh and the eighth ones which will be discussed later. Their features can efficiently be computed.

\begin{Lemma}
Given a meta-path $\mathbf{P} = \mathbf{P}_1 \times \mathbf{P}_2^T$ where $\mathbf{P}_1$ is a simple meta-path, the computation of Eq.~(\ref{eq:relfea1}) on $\mathbf{P}$ can be reduced to:
\begin{align}
\mathbf{z} = \mathbf{P}_1 \times (\mathbf{D}_2 \times \mathbf{P}_2^T \times \mathbf{y})
\label{eq:relfea2}
\end{align}
where $\mathbf{D}_2 \in \mathbb{R}^{n_t \times n_t}$ is a diagonal matrix and $\mathbf{D}_2[i,i] = 1/\sum_j \mathbf{P}_2[j,i]$.
\label{le1}
\end{Lemma}
\begin{Proof}
Because $\mathbf{P}_1$ is a simple meta-path, {\em i.e.}, a many-to-one relation, there is only one value per row, {\em i.e.}, $\mathbf{P}_1[i,k_i]=1$ and $\mathbf{P}_1[i,j]=0,~\forall j \ne k_i$.
\begin{align}
1/\mathbf{D}[i,i]
& = \sum_j \mathbf{P}[i,j] \nonumber \\
& = \sum_j \sum_k \mathbf{P}_1[i,k] \mathbf{P}_2^T[k,j] \nonumber \\
& = \sum_j \sum_k \mathbf{P}_1[i,k] \mathbf{P}_2[j,k] \nonumber \\
& = \sum_j \mathbf{P}_2[j,k_i] \nonumber \\
& = 1/\mathbf{D}_2[k_i,k_i]
\end{align}
Assume $\mathbf{T} = \mathbf{D}^{-1} \times \mathbf{P}_1 \times \mathbf{D}_2$.
\begin{align}
\mathbf{T}[i,j]
& = 1/\mathbf{D}[i,i] \times \mathbf{P}_1[i,j] \times \mathbf{D}_2[j,j] \nonumber \\
& = \left\{
\begin{array}{ll}
    1,&\text{if}~j=k_i\\
    0,&\text{otherwise}
\end{array}
\right.
\end{align}
Hence, $\mathbf{T} = \mathbf{P}_1$, $\mathbf{D} \times \mathbf{P}_1 = \mathbf{P}_1 \times \mathbf{D}_2$, and thus Eq.~(\ref{eq:relfea1}) is equivalent to Eq.~(\ref{eq:relfea2}).
\end{Proof}

Eq.~(\ref{eq:relfea2}) is important because it enables us to obtain the weighted label fraction of linked nodes without extensive matrix operations. In this manner, we can effectively avoid computing the redundant full matrix of a meta-path as an intermediate result.

\subsubsection{Many-to-Many Cases}

Let's consider the meta-path ``transaction $\xrightarrow{containsItem}$ item $\xrightarrow{isTitle}$ title $\xrightarrow{isTitle^{-1}}$ item $\xrightarrow{containsItem^{-1}}$ transaction" in Table~\ref{tab:metapath}. It involves a many-to-many link type, {\em i.e.}, ``transaction $\xrightarrow{containsItem}$ item", in Figure~\ref{fig:schema}.

\begin{Definition}[Complex Meta-path]
A meta-path is a complex meta-path if it contains at least one many-to-many relation.
\end{Definition}

Note that a complex meta-path is ``usually" a (composite) many-to-many relation. Now, we dichotomize all meta-paths into simple meta-paths and complex meta-paths. Given any nontrivial meta-path ($l>1$), we still decompose it into two parts at the node type of the smallest size, $\mathbf{P} = \mathbf{P}_1 \times \mathbf{P}_2^T$, and we propose to compute its features as follows:
\begin{align}
\mathbf{z} = \mathbf{D}_1 \times \mathbf{P}_1 \times (\mathbf{D}_2 \times \mathbf{P}_2^T \times \mathbf{y})
\label{eq:relfea3}
\end{align}
where $\mathbf{D}_1 \in \mathbb{R}^{n \times n}$ is a diagonal matrix and $\mathbf{D}_1[i,i] = 1/\sum_j \mathbf{P}_1[i,j]$. Obviously, $\mathbf{D}_1 = \mathbf{I}_n$ where $\mathbf{I}_n$ is an identity matrix when $\mathbf{P}_1$ is a simple meta-path. Therefore, Eq.~(\ref{eq:relfea2}) is a special case of Eq.~(\ref{eq:relfea3}) in the many-to-one scenarios.

However, Eq.~(\ref{eq:relfea3}) is not equivalent to Eq.~(\ref{eq:relfea1}) or Eq.~(\ref{eq:relfea2}) when $\mathbf{P}_1$ is a complex meta-path. For example, $\mathcal{P}=$``transaction $\xrightarrow{containsItem}$ item $\xrightarrow{isTitle}$ title $\xrightarrow{isTitle^{-1}}$ item $\xrightarrow{containsItem^{-1}}$ transaction" can be decomposed as $\mathcal{P}_1=\mathcal{P}_2=$``transaction $\xrightarrow{containsItem}$ item $\xrightarrow{isTitle}$ title" where $\mathcal{P}_1$ is a complex meta-path. In Eq.~(\ref{eq:relfea1}), we compute the average fraction of frauds in the linked transactions and each transaction is weighted by the number of common game titles in the current transaction. It does not, however, distinguish which game titles are shared. In other words, each shared game title is counted equally. In contrast, Eq.~(\ref{eq:relfea3}) counts a shared rare title more than a shared popular title because $\mathbf{D}_2$ accounts for a normalization step at the title level. The similarity between two transactions increases proportionally to the number of titles they share, but is offset by the popularity of the title.

Here, we present an analogy to term frequency-inverse document frequency (tf-idf) in information retrieval. Essentially, Eq.~(\ref{eq:relfea1}) computes the similarity between linked transactions in a term frequency manner, while Eq.~(\ref{eq:relfea3}) captures the tf-idf information. Most importantly, Eq.~(\ref{eq:relfea1}) is intractable when $\mathbf{P} \in \mathbb{R}^{n \times n}$ is too large and nearly full, while Eq.~(\ref{eq:relfea3}) only needs to deal with sparse matrices $\mathbf{P}_1,\mathbf{P}_2 \in \mathbb{R}^{n \times n_t}$ where $n_t$ is typically much smaller than $n$.

\subsection{Collective Fraud Detection}

So far we have explored the inter-transaction dependency to capture the correlated fraudulent behaviors where the correlations mainly exist between training transactions and test transactions. We further notice that fraudulent behaviors are fast evolving. For example, a batch of new transactions may be made by the same new billing account but with different IP addresses, some of which are rather risky and others might be proxy. It is desirable to mark all these transactions as suspicious.

As stated in Section~\ref{sec:problem}, the inference problem for collective fraud detection in a HIN is to learn a predictive function $f:(\mathcal{V},\mathcal{E},\mathbf{X})\to\mathbf{y}$. Conventional classification approaches usually make an independent and identically distributed (i.i.d.) assumption, and thus the probability of each transaction being frauds is inferred independently as $f(\mathbf{x}_i)\propto\text{Pr}(y_i=1|\mathbf{x}_i)$. In addition to the given features $\mathbf{X}$, we include features $\big\{\mathbf{z}^1,\cdots,\mathbf{z}^c\big\}$ that are derived from meta-paths $\big\{\mathcal{P}^1,\cdots,\mathcal{P}^c\big\}$ where $c$ is the number of extracted meta-paths. Therefore, the target is to learn
\begin{align}
f(\mathbf{x}_i)\propto\text{Pr}(y_i=1|\mathbf{x}_i,z_i^1,\cdots,z_i^c)
\label{eq:obj}
\end{align}

\begin{figure}[t]
\centering
    \begin{minipage}[l]{1.0\columnwidth}
      \centering
      \includegraphics[width=1\textwidth]{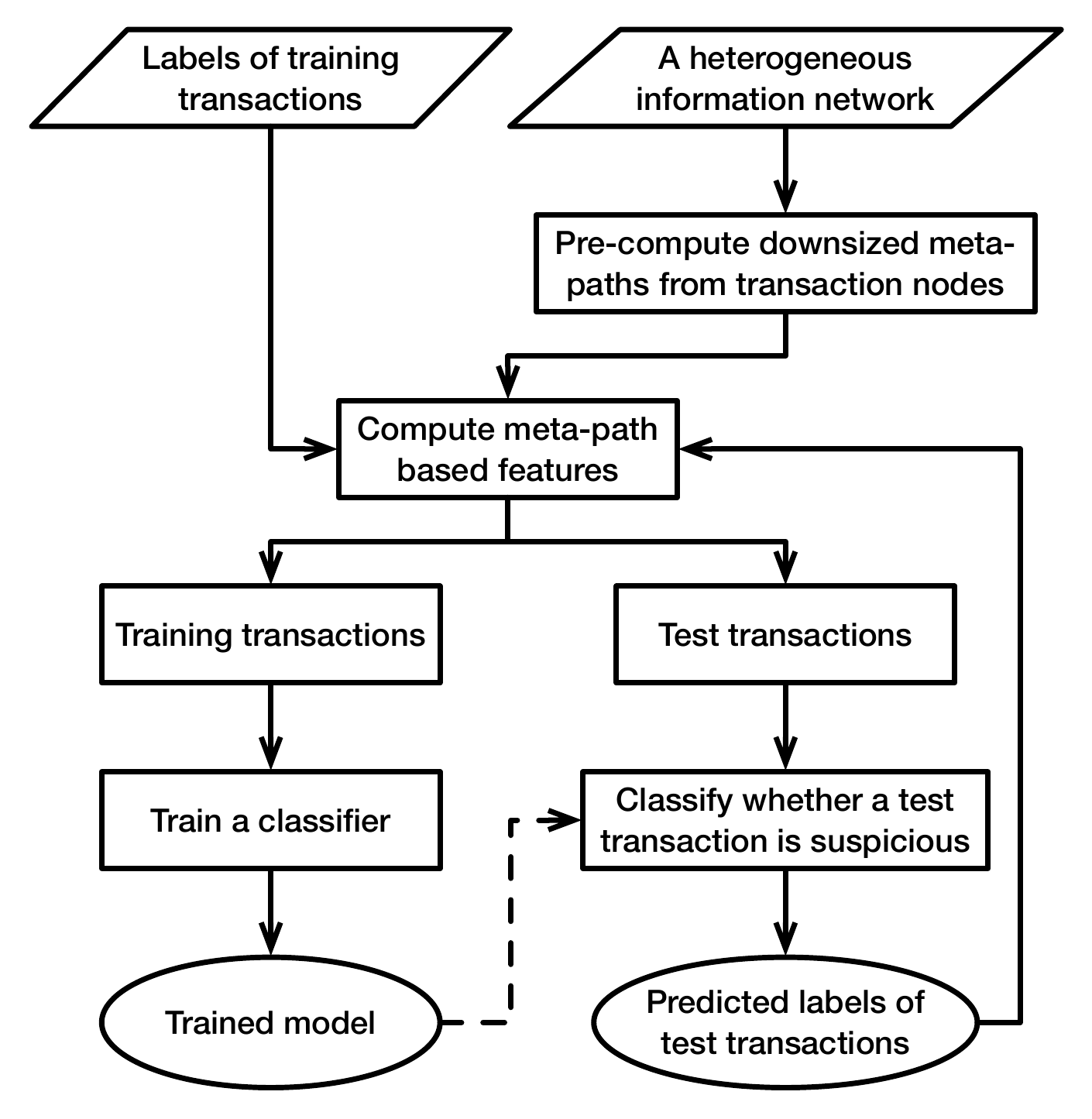}
    \end{minipage}
  \caption{The flowchart of \pro.}\label{fig:flowchart}
\end{figure}

\begin{algorithm}[t]
\caption{\pro}
\label{algo:pro}
\begin{algorithmic}[1]
\REQUIRE\texttt{$\mathcal{G}$: a HIN, $\mathbf{X}$: a feature matrix, $\mathbf{y}$[train]: labels of training instances, $\mathcal{A}$: a base classifier}
\ENSURE\texttt{$\mathbf{y}$[test]: labels of test instances}
\STATE\texttt{/* pre-compute downsized meta-paths */}
\STATE\texttt{p=q=0, paths[0]=$\mathbf{I}_n$, traces[0]=[1]}
\REPEAT
\FOR{\texttt{$\mathcal{E}^k$=<$\mathcal{V}^i$,$\mathcal{V}^j$>$\in\mathcal{E}$}}
\IF{\texttt{traces[p][end]=i} \AND \texttt{|$\mathcal{V}^i$|>|$\mathcal{V}^j$|}}
\STATE\texttt{paths[++q]=paths[p]$\times\mathbf{A}^k$}
\STATE\texttt{traces[q]=[traces[p],j]}
\ENDIF
\ENDFOR
\STATE\texttt{p++}
\UNTIL{\texttt{p>q}}
\STATE\texttt{/* collective prediction */}
\STATE\texttt{f=$\mathcal{A}$($\mathbf{X}$[train,:],$\mathbf{y}$[train])}
\STATE\texttt{$\mathbf{y}$[test]=f($\mathbf{X}$[test,:])}
\REPEAT
\STATE\texttt{c=0}
\FOR{\texttt{$\mathbf{P}_1$=paths[i]$\in$paths}}
\FOR{\texttt{$\mathbf{P}_2$=paths[j]$\in$paths}}
\IF{\texttt{traces[i][end]=traces[j][end]}}
\STATE\texttt{compute meta-path features $\mathbf{z}^{++c}$ by Eq.~(\ref{eq:relfea3})}
\ENDIF
\ENDFOR
\ENDFOR
\STATE\texttt{$\mathbf{X}'$=[$\mathbf{X}$,$\mathbf{z}^1$,$\cdots$,$\mathbf{z}^c$]}
\STATE\texttt{f=$\mathcal{A}$($\mathbf{X}'$[train,:],$\mathbf{y}$[train])}
\STATE\texttt{$\mathbf{y}$[test]=f($\mathbf{X}'$[test,:])}
\UNTIL{\texttt{convergence}}
\end{algorithmic}
\end{algorithm}

In this manner, however, the inference of different transactions is essentially not independent, because meta-path based features contain the label information of linked transactions in both training and test sets. It can be done in an iterative framework where the label of a transaction is inferred based on the labels of its linked transactions through its meta-path based features, and its predicted label will further be used to infer the labels of its linked transactions by updating their meta-path based features. It is similar to the framework of Heterogeneous Collective Classification (HCC) \cite{kong2012meta}, and we improve it with a much more efficient way of computing meta-path based features.

\begin{Definition}[Downsized Meta-path]
Given the node sequence of a meta-path $\mathcal{P}=<\mathcal{V}^{k_0},\cdots,\mathcal{V}^{k_l}>$, it is a downsized meta-path if $n_{k_0}>\cdots>n_{k_l}$.
\end{Definition}

We design the process of meta-path exploration based on discussions in the last section which can be summarized into two facts: (1) Meta-paths that are used for feature computation in our task always start from and end at transaction nodes. (2) Each meta-path can be decomposed into two parts at the node type of the smallest size. Therefore, we can perform a breadth-first-search from transaction nodes to find all downsized meta-paths from $\mathcal{V}^1$ to each other node type. In the search procedure, say, the current meta-path $\mathbf{P}$ is from $\mathcal{V}^1$ to $\mathcal{V}^i$, we enumerate link types ``$\mathcal{V}^i \xrightarrow{\mathcal{E}^k} \mathcal{V}^j$" in the network schema. If $|\mathcal{V}^i|>|\mathcal{V}^j|$, a new meta-path $\mathbf{P}' = \mathbf{P} \times \mathbf{A}^k$ is added into $\mathcal{S}_j$. Search will be expanded from the newly added meta-paths until all downsized meta-paths from $\mathcal{V}^1$ to $\mathcal{V}^i$ have been found and included in $\mathcal{S}_i$.

In contrast to the original HCC framework where all meta-paths connecting the target entities are pre-computed in the initialization step which results in a lot of redundancy as discussed in Section~\ref{sec:redundancy}, we only pre-compute downsized meta-paths and then organically combine them using Eq.~(\ref{eq:relfea3}) in the feature computation step. In this manner, we have two bonus effects: (1) Redundant meta-paths are automatically avoided because a concatenation of two downsized meta-paths is for sure not a sub-sequence of another one. (2) There is no more need to manually tune the maximal length of meta-paths.

The flowchart of the proposed method {\pro} is outlined in Figure~\ref{fig:flowchart} where the predicted labels of test transactions are initialized from a classifier that is trained using the given features $\mathbf{X}$ only. Then a new classifier is trained using additional meta-path based features which are iteratively updated based on both the labels of training transactions and the predicted labels of test transactions. The pseudo-code is presented in Algorithm~\ref{algo:pro}.

\section{Experiments}
\label{sec:exp}

\begin{figure*}[t]
\centering
    \begin{minipage}[l]{2.0\columnwidth}
      \centering
      \includegraphics[width=1\textwidth]{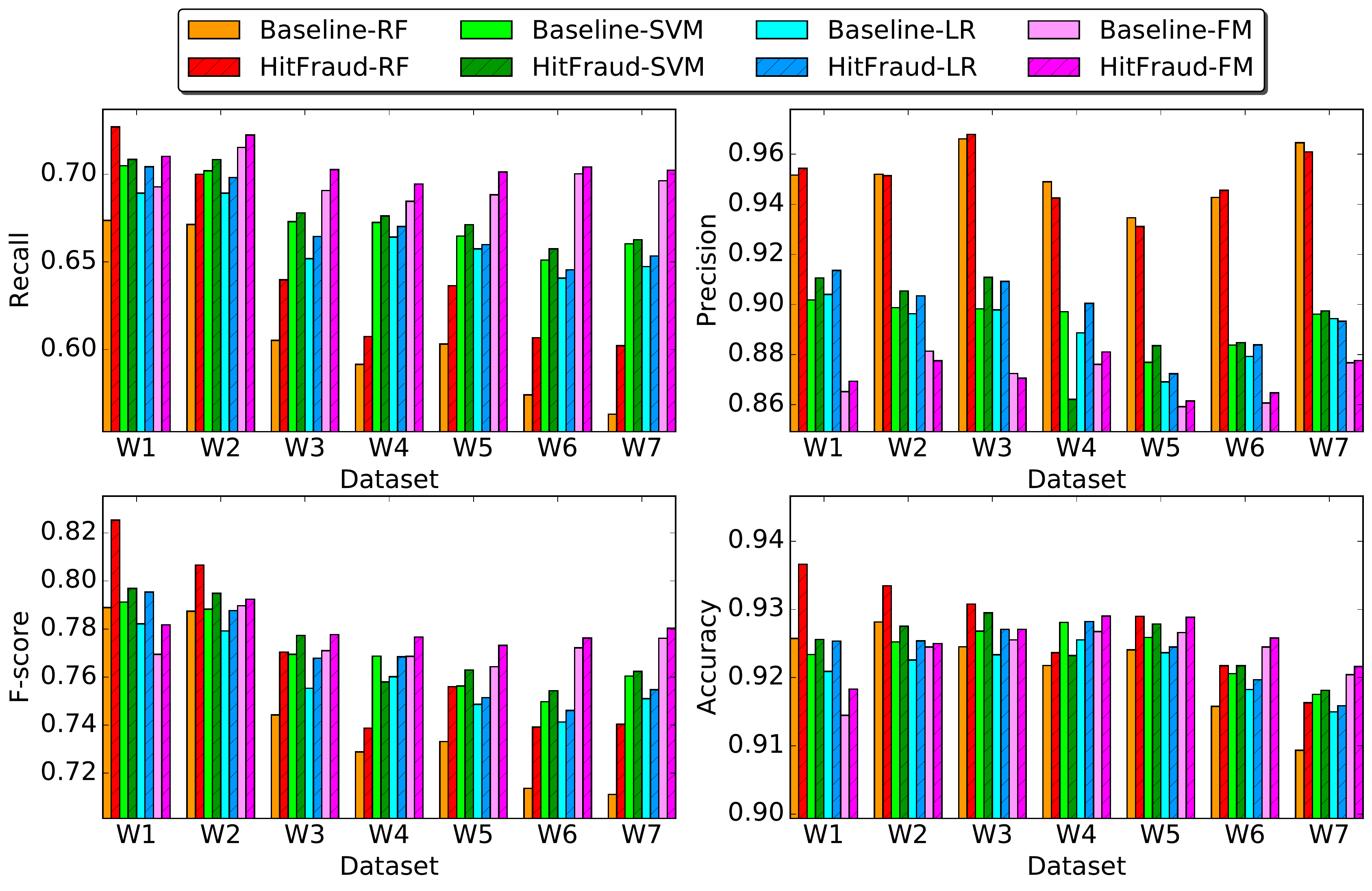}
    \end{minipage}
  \caption{Prediction performance.}\label{fig:performance}
\end{figure*}

\begin{figure*}[t]
\centering
    \begin{minipage}[l]{2.0\columnwidth}
      \centering
      \includegraphics[width=1\textwidth]{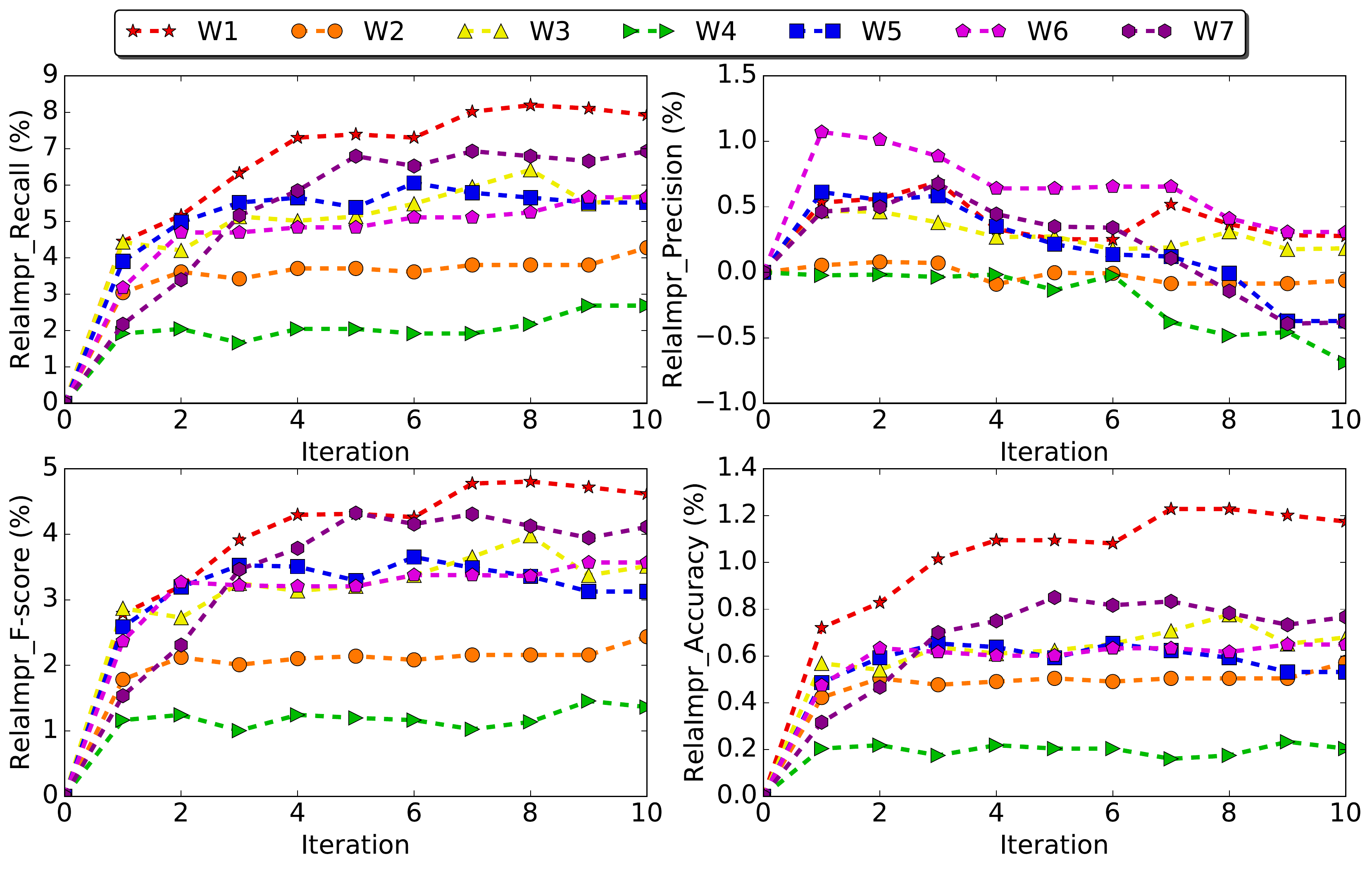}
    \end{minipage}
  \caption{Convergence analysis of {\pro} with random forest as the base classifier.}\label{fig:convergence_rf}
\end{figure*}

\subsection{Experimental Setup}

As introduced in Section~\ref{sec:dataset}, experiments are conducted on {\abbrname} payment transaction data that contains $n=130K$ transactions with manual review labels. For training and evaluation purposes, we segment the data into two consecutive parts so that one-week data is used for testing models and the preceding weeks are used for training. Based on sliding windows, 7 data segmentations are created, each of which is denoted as {\em W1} to {\em W7}. Four metrics are reported: recall, precision, F-score and accuracy.

\subsection{Effectiveness with Various Classifiers}

One claim of this paper is that {\pro} can work well in conjunction with a variety of base classifiers. To evaluate this claim, we conduct experiments using various base classifiers, including random forest (RF), support vector machines (SVM), logistic regression (LR), and factorization machines (FM). The implementations of these base classifiers from \texttt{scikit-learn}\footnote{\url{http://scikit-learn.org}} and \texttt{fastFM}\footnote{\url{https://github.com/ibayer/fastFM}} are used with default hyperparameter configurations. The baselines are the same base classifiers that explore only the given feature space $\mathbf{X}$. Figure~\ref{fig:performance} shows the prediction performance comparing {\pro} to the baselines. We can observe that, by conducting collective fraud detection, {\pro} is usually able to outperform the baselines with multiple choices of base classifiers, on multiple datasets, in multiple evaluation metrics. For example, with random forest on the {\em W1} dataset, recall is boosted 7.93\% from 0.6737 to 0.7271 ($p=0.000$), precision 0.28\% from 0.9517 to 0.9543 ($p=0.218$), F-score 4.62\% from 0.7889 to 0.8253 ($p=0.000$), and accuracy 1.17\% from 0.9257 to 0.9366 ($p=0.003$). Recall, F-score and accuracy are all significant, although precision is non-significant. In few cases, precision is sacrificed to boost recall and the overall F-score. However, precision and recall can be improved at the same time by {\pro} in most cases. In general, it demonstrates that {\pro} is flexible and effective in conjunction with a diversity of underlying classification algorithms, and the inter-transaction dependency can indeed be explored to more effectively capture fraud payment transactions.

Moreover, note that $c=38$ meta-paths linking transactions are obtained after pairwise combination at the target node of $17$ downsized meta-paths that are extracted from the network schema in Figure~\ref{fig:schema}. It is impressive to see such an improvement from these $c=38$ meta-path based features compared to the baselines that use the given features of $d=2K$ dimensions. A side observation is that random forest typically achieves the best precision while factorization machines can usually obtain the best recall. For practical uses, an ensemble model is desirable on the top of {\pro} for the best performance of fraud detection.

\subsection{Fast Convergence}

The number of iterations for collective prediction is set to $10$ in the results reported above. Another claim of this paper is that {\pro} can actually converge fast. In order to verify this claim, Figure~\ref{fig:convergence_rf} shows the detailed changes of evaluation metrics on the predictions of test transactions during the iterative process. Due to space limit, we present the convergence analysis of random forest only, and similar observations can be found on other base classifiers. Note that the predicted labels of test transactions are initialized from the baseline in Algorithm~\ref{algo:pro}, therefore we denote the baseline performance as the value at iteration 0. Since the absolute prediction performances are different across datasets, here we plot the relative improvement of {\pro} over the baseline which has the following mathematical form
\begin{align}
RelaImpr=\left(\frac{metric(\pro)}{metric(baseline)}-1\right) \times 100\%
\end{align}

It can be observed that recall, F-score and accuracy are significantly improved in the first few iterations, and they become stable in the late stages. In terms of precision, it is slightly degraded in some datasets. However, we consider this as a valuable trade-off, because we use no more than 1\% decrease in precision to exchange for nearly 8\% increase in recall, and thereby always improve F-score and accuracy. More importantly, recall is more critical in practice when the predicted labels are used as assistance for manual review of suspicious transactions.

\begin{figure}[t]
\centering
    \begin{minipage}[l]{1.0\columnwidth}
      \centering
      \includegraphics[width=1\textwidth]{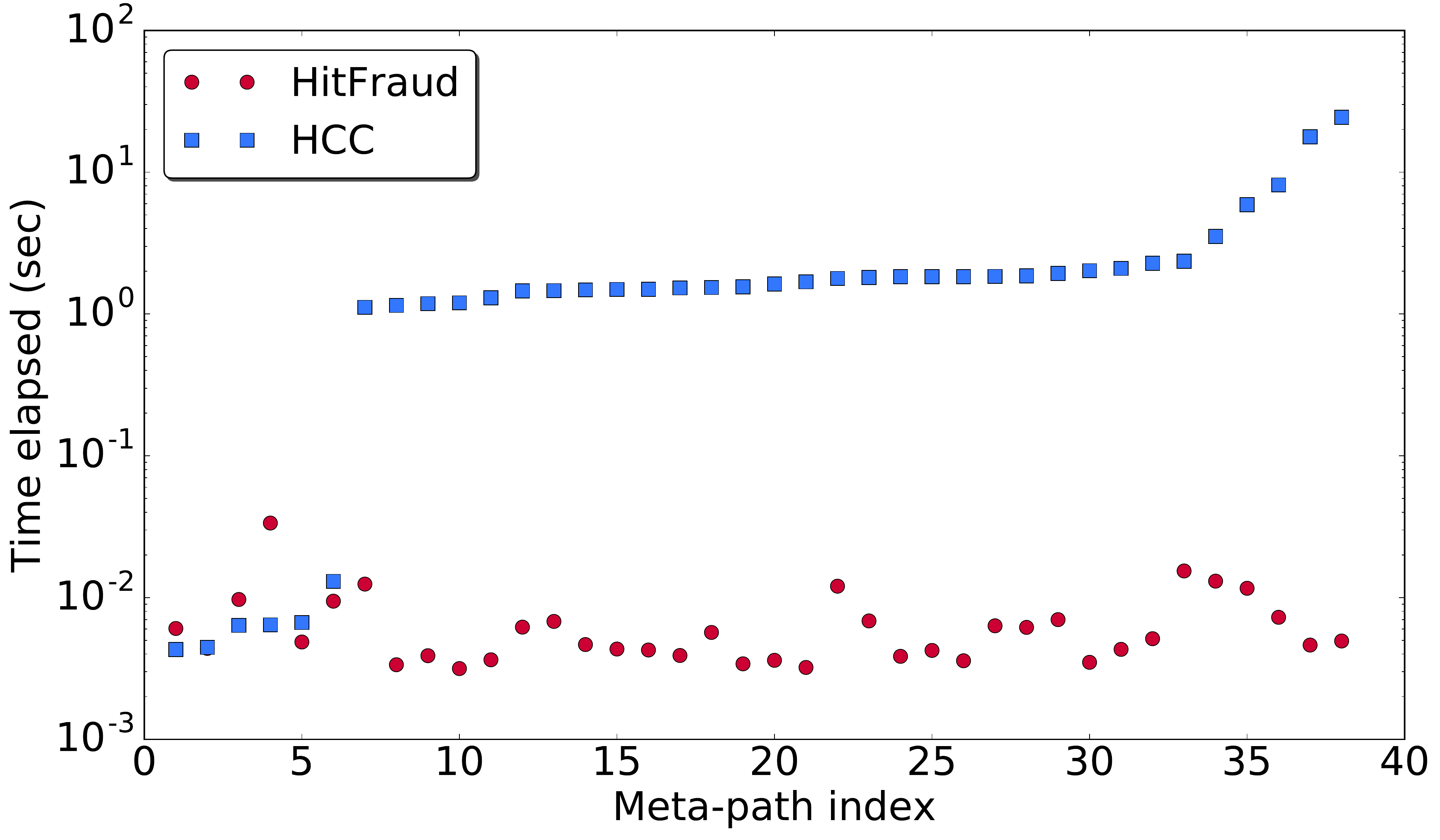}
    \end{minipage}
  \caption{Computation cost of meta-paths.}\label{fig:time}
\end{figure}

\begin{table*}[t]
\caption{Summary of related work on graph based fraud/anomaly/outlier/spam detection.}
\label{tab:related}
\centering
\begin{tabular}{||l|l|l||}
\hline
Method & Input & Output \\
\hline
\hline
{\pro} & a heterogeneous graph ({\abbrname} payment transaction graph) & nodes (transactions) \\
ABCOutliers \cite{gupta2013detecting} & a heterogeneous graph (Wikipedia entity graph) & subgraphs (entity groups) \\
StreamSpot \cite{manzoor2016fast} & multiple heterogeneous graphs (information flow graphs) & graphs (system logs) \\
CloseMine \cite{liu2005mining} & multiple heterogeneous graphs (software behavior graphs) & graphs (program runs) \\
FRAUDAR \cite{hooi2016fraudar} & a bipartite graph (social graph) & nodes (users) \\
fBox \cite{shah2014spotting} & a bipartite graph (social, rating graph) & nodes (users) \\
CopyCatch \cite{beutel2013copycatch} & a bipartite graph (Page Likes graph) & nodes (users) \\
FocusCO \cite{perozzi2014focused} & a homogeneous graph (DBLP, co-purchase, citation graph) & nodes (authors/movies/bloggers) \\
CatchSync \cite{jiang2014catchsync} & a homogeneous graph (social graph) & nodes (users) \\
SybilRank \cite{cao2012aiding} & a homogeneous graph (social graph) & nodes (users) \\
Collusionrank \cite{ghosh2012understanding} & a homogeneous graph (social graph) & nodes (users) \\
CODA \cite{gao2010community} & a homogeneous graph (DBLP graph) & nodes (conferences/authors) \\
TrustRank \cite{gyongyi2004combating} & a homogeneous graph (web graph) & nodes (pages) \\
AMEN \cite{perozzi2016scalable} & a homogeneous graph (DBLP, social graph) & subgraphs (author/user groups) \\
SODA \cite{gupta2014local} & a homogeneous graph (DBLP, yeast graph) & subgraphs (author/protein groups) \\
\hline
\end{tabular}
\end{table*}

\subsection{Efficiency in Meta-Path Computation}

As we discussed in Section~\ref{sec:redundancy}, there is much redundancy in the plain-vanilla computation of meta-path based features which aggravates not only the time cost but also the memory consumption. There are in total $c=38$ meta-paths explored in this work, and Figure~\ref{fig:time} compares the time cost of computing each meta-path based feature between the approach proposed in Section~\ref{sec:computation} for {\pro} and HCC presented in \cite{kong2012meta}. The comparison of memory cost is omitted due to space limit, and the trend is similar to Figure~\ref{fig:time}. Note that this experiment is conducted on a network constructed from one-week data, because the time and space cost for HCC is formidable on the whole dataset. We can observe that the discrepancy of time cost is significant even on the log scale. For a few cases where {\pro} and HCC take almost the same time, those are meta-paths that involve nearly one-to-one relation where redundancy is not very severe, {\em e.g.}, ``transaction $\xrightarrow{byUser}$ user $\xrightarrow{byUser^{-1}}$ transaction". Obviously, it is not very likely that many users would place multiple transactions within one week.

\subsection{Case Study}

We investigate the statistical significance of each meta-path based feature in identifying suspicious transactions. On the {\em W1} dataset, we first obtain the meta-path based feature values after convergence using random forest as the base classifier, and then randomly sample 1,000 transactions from the test set. Welch's t-test is performed on the fraud group and the normal group for each meta-path. It is found that each feature associated with the meta-paths in Table~\ref{tab:metapath} can effectively distinguish suspicious transactions from normal ones with significance level $\alpha$ = 0.05. From the results, we can see that the correlated and fast evolving fraudulent behaviors from the shared billing accounts, shared IP addresses, shared user accounts, and shared game titles all play very significant roles in determining the fraud likelihood of a test transaction.



\section{Related Work}
\label{sec:related}

This work is related to both fraud detection and collective classification techniques. We briefly discuss both of them. Conventional fraud detection methods focus on detecting deceptive opinion spams through review texts which, however, can be evaded by intelligent adversaries such as carefully selecting words in a review \cite{jindal2008opinion,ott2011finding}. Graph based methods detect frauds by leveraging the linkage information between entities of interest which are relatively harder to evade because making a fraud transaction unavoidably generates links in the graph \cite{noble2003graph,eberle2007discovering,hooi2016fraudar}. However, the related work on graph based fraud detection summarized in Table~\ref{tab:related} are mostly based on either homogeneous information networks or bipartite graphs, and those for heterogeneous information networks aim to find frauds at the level of graphs or subgraphs. In contrast, we study the fraud detection problem at a fine-grained level of nodes (transactions) in a complex and semantically enriched heterogeneous information network.


Collective classification is to predict the labels for a group of related instances simultaneously, rather than predicting a label for each instance independently. In relational datasets, the label of one instance can be related to the labels of other related instances. A local classifier is usually employed to iteratively classify unlabeled instances using both features of the instances and relational features derived from related instances \cite{sen2008collective}. It involves an iterative process to update the labels and the relational features of related instances. Many classifiers have been used, including logistic regression \cite{lu2003link}, na\"{\i}ve Bayes \cite{neville2000iterative}, relational dependency network \cite{neville2003collective}, {\em etc.} The concept of meta-paths has been applied on mining P2P lending networks \cite{zhang2016netcycle}, social networks \cite{cao2015inferring}, bioinformatic networks \cite{cao2014collective,kong2013multi}, and bibliographic networks \cite{kong2012meta,sun2011pathsim}. However, these approaches are not scalable because the concrete form of a meta-path is needed.

\section{Conclusion}
\label{sec:conclusion}

In this paper, we propose {\pro}, a collective fraud detection algorithm that captures the inter-transaction dependency. Meta-path based features are efficiently computed through the use of pre-computing downsized meta-paths. Suspicious transactions in the test set are collectively identified when they share common fraudulent behaviors. Experiments on {\abbrname} payment transaction data demonstrate that the prediction performance is effectively boosted by {\pro} with different choices of base classifiers and with fast convergence. It is validated that the correlated and fast evolving fraudulent behaviors can indeed be explored to more effectively capture fraud payment transactions.

\bibliographystyle{plain}
\bibliography{reference}

\end{document}